%% file: neurips_2024.tex
\documentclass{article}

\usepackage[final,nonatbib]{_sty/neurips_2024}

\usepackage[utf8]{inputenc} 
\usepackage[T1]{fontenc}    
\usepackage{hyperref}       
\usepackage{url}            
\usepackage{booktabs}       
\usepackage{amsfonts}       
\usepackage{nicefrac}       
\usepackage{microtype}      
\usepackage{xcolor}         
\usepackage{comment}
\usepackage{colortbl}
\usepackage[square,numbers,sort]{natbib}
\usepackage{graphicx}
\usepackage{amsmath}
\usepackage{multirow}
\usepackage{wrapfig}
\definecolor{mygray}{gray}{.9}
\usepackage{subcaption}

\title{Infinite-Dimensional Feature Interaction}

\author{%
Chenhui Xu$^{1,2}$\;  \hspace{3mm} Fuxun Yu$^{1,3}$\; \hspace{3mm} Maoliang Li$^4$\; \hspace{3mm} Zihao Zheng$^4$\\\textbf{ Zirui Xu$^{1}$\; \hspace{3mm} Jinjun Xiong$^{2,}$\thanks{Corresponding Author.}\; \hspace{3mm} Xiang Chen$^{{1,4,}}\footnotemark[1]$}\\
$^1$George Mason University
  \hspace{10mm}$^2$University at Buffalo
\\ $^3$Microsoft\; \hspace{15mm}$^4$Peking University\; \\
\texttt{\{cxu26,jinjun\}@buffalo.edu, xiang.chen@pku.edu.cn}
}

\begin{document}

\maketitle

\input{_txt/0_Abstract}
\input{_txt/1_Introduction}
\input{_txt/2_Related_Work}
\input{_txt/3_SpaceTransfer}

\input{_txt/4_InfiNet}

\input{_txt/5_Experiments}
\input{_txt/6_Conclusion}

\newpage
{\small

\bibliographystyle{abbrv}
\bibliography{_bib/ref}
}

\newpage
\appendix

\input{_txt/8_Appendix}


\newpage
\newpage
\section*{NeurIPS Paper Checklist}

\begin{enumerate}

\item {\bf Claims}
    \item[] Question: Do the main claims made in the abstract and introduction accurately reflect the paper's contributions and scope?
    \item[] Answer: \answerYes{} 
    \item[] Justification: We attribute the success of modern models to an feature interaction and claim the kernel methods can expand this feature interaction space. And propose the InfiNet. They are included in the abstract and introduction.

\item {\bf Limitations}
    \item[] Question: Does the paper discuss the limitations of the work performed by the authors?
    \item[] Answer:  \answerYes{}
    \item[] Justification: After the Conclusion

\item {\bf Theory Assumptions and Proofs}
    \item[] Question: For each theoretical result, does the paper provide the full set of assumptions and a complete (and correct) proof?
    \item[] Answer: \answerYes{} 
    \item[] Justification: In Section~\ref{sec:3} and \ref{sec:4}.

    \item {\bf Experimental Result Reproducibility}
    \item[] Question: Does the paper fully disclose all the information needed to reproduce the main experimental results of the paper to the extent that it affects the main claims and/or conclusions of the paper (regardless of whether the code and data are provided or not)?
    \item[] Answer: \answerYes{} 
    \item[] Justification: In Section~\ref{sec:6} and Appendix \ref{app:training}.

\item {\bf Open access to data and code}
    \item[] Question: Does the paper provide open access to the data and code, with sufficient instructions to faithfully reproduce the main experimental results, as described in supplemental material?
    \item[] Answer: \answerYes{} 
    \item[] Justification: All datasets we use in this paper are third-party open source.

\item {\bf Experimental Setting/Details}
    \item[] Question: Does the paper specify all the training and test details (e.g., data splits, hyperparameters, how they were chosen, type of optimizer, etc.) necessary to understand the results?
   \item[] Answer: \answerYes{} 
    \item[] Justification: In Section~\ref{sec:6} and Appendix \ref{app:training}.

\item {\bf Experiment Statistical Significance}
    \item[] Question: Does the paper report error bars suitably and correctly defined or other appropriate information about the statistical significance of the experiments?
    \item[] Answer: \answerNo{} 
    \item[] Justification: Training for multiple times is too expensive.

\item {\bf Experiments Compute Resources}
    \item[] Question: For each experiment, does the paper provide sufficient information on the computer resources (type of compute workers, memory, time of execution) needed to reproduce the experiments?
    \item[] Answer: \answerYes{} 
    \item[] Justification: In Section \ref{sec:6}.

\item {\bf Code Of Ethics}
    \item[] Question: Does the research conducted in the paper conform, in every respect, with the NeurIPS Code of Ethics \url{https://neurips.cc/public/EthicsGuidelines}?
    \item[] Answer: \answerYes{} 
    \item[] Justification: This paper is with the NeurIPS Code of Ethics.

\item {\bf Broader Impacts}
    \item[] Question: Does the paper discuss both potential positive societal impacts and negative societal impacts of the work performed?
    \item[] Answer: \answerYes{} 
    \item[] Justification: After the Conclusion

\item {\bf Safeguards}
    \item[] Question: Does the paper describe safeguards that have been put in place for responsible release of data or models that have a high risk for misuse (e.g., pretrained language models, image generators, or scraped datasets)?
    \item[] Answer: \answerNA{} 
    \item[] Justification: No such risks.

\item {\bf Licenses for existing assets}
    \item[] Question: Are the creators or original owners of assets (e.g., code, data, models), used in the paper, properly credited and are the license and terms of use explicitly mentioned and properly respected?
    \item[] Answer: \answerYes{} 
    \item[] Justification: All datasets and baselines are cited.

\item {\bf New Assets}
    \item[] Question: Are new assets introduced in the paper well documented and is the documentation provided alongside the assets?
    \item[] Answer: \answerNA{} 
    \item[] Justification: No new assets.

\item {\bf Crowdsourcing and Research with Human Subjects}
    \item[] Question: For crowdsourcing experiments and research with human subjects, does the paper include the full text of instructions given to participants and screenshots, if applicable, as well as details about compensation (if any)?
    \item[] Answer: \answerNA{} 
    \item[] Justification: The paper does not involve crowdsourcing nor research with human subjects

\item {\bf Institutional Review Board (IRB) Approvals or Equivalent for Research with Human Subjects}
    \item[] Question: Does the paper describe potential risks incurred by study participants, whether such risks were disclosed to the subjects, and whether Institutional Review Board (IRB) approvals (or an equivalent approval/review based on the requirements of your country or institution) were obtained?
    \item[] Answer: \answerNA{} 
    \item[] Justification: The paper does not involve crowdsourcing nor research with human subjects.

\end{enumerate}

\end{document}

%% file: _txt/0_Abstract.tex
\begin{abstract}
The past neural network design has largely focused on feature \textit{representation space} dimension and its capacity scaling (e.g., width, depth), but overlooked the feature \textit{interaction space} scaling. 
 Recent advancements have shown shifted focus towards element-wise multiplication to facilitate higher-dimensional feature interaction space for better information transformation. Despite this progress, multiplications predominantly capture low-order interactions, thus remaining confined to a finite-dimensional interaction space. To transcend this limitation, classic kernel methods emerge as a promising solution to engage features in an infinite-dimensional space. We introduce InfiNet, a model architecture that enables feature interaction within an infinite-dimensional space created by RBF kernel. Our experiments reveal that InfiNet achieves new state-of-the-art, owing to its capability to leverage infinite-dimensional interactions, significantly enhancing model performance.
\end{abstract}

%% file: _txt/1_Introduction.tex
\section{Introduction}
\begin{figure}[t]    \centering
    \includegraphics{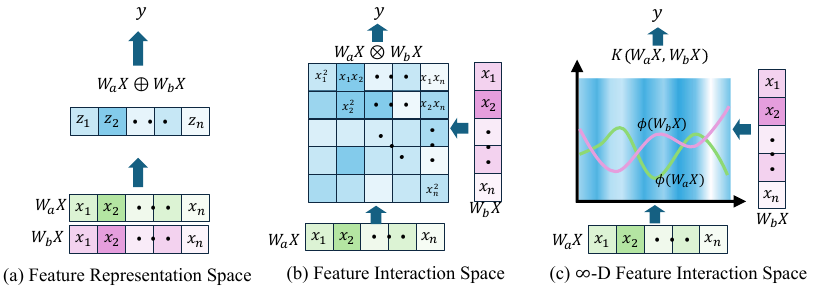}
    \caption{(a) Traditional feature representation without interaction~\cite{woo2023convnext,he2016deep}. (b) Recent work with finite feature interaction~\cite{rao2022hornet,xu2022quadralib}. (c) Our method: Kernel-enabled infinite feature interaction.}
    \label{fig:1}
    \vspace{-5mm}
\end{figure}

In the past decade, deep neural network architecture design has experienced several major paradigm shifts regarding the feature representation learning.
As shown in Fig.~\ref{fig:1}(a), the early stage of neural network design is dominant by flat stream architectures in the form of \textit{weight-feature interaction} (e.g., $W \textbf x$), like multi-layer perceptron (MLP), convolution neural networks (CNN), ResNet, etc.
These models usually adopt linear superposition (e.g., $W_i \textbf x \oplus W_j \textbf x$)\footnote[1]{The symbol $\oplus$ denotes the Direct Sum of vector spaces. This corresponds to structures such as channel expansion/bottleneck in commonly used neural networks. While the $\otimes$ is elementwise multiplication.} in the feature representation space.
Therefore, the feature representation space scaling is limited to increase model channel width and depth~\cite{krizhevsky2012imagenet,he2016deep}.
Nevertheless, this scaling approach has witnessed model development from the very small-scale MLPs or LeNet\cite{lecun1998gradient} to the recent huge ConvNext V2~\cite{woo2023convnext}. 
With the ultra-scaled parameter amounts, computing complexity, and model size, the return of investment on model performance by further scaling feature dimensions has largely plateaued~\cite{dohmatob2024tale}.



Despite the plateau in feature representation space, recent sporadic architecture design works~\cite{ma2024rewrite} shed light on another potential dimension of scaling: feature interaction space.
Specifically, as shown in Fig.~\ref{fig:1}(b), these neural network designs generally demonstrate \textit{feature-feature interaction} (e.g., $W_i \textbf x \otimes W_j \textbf x$).
As a mathematical example, the self-attention mechanism in Transformers~\cite{vaswani2017attention} can be formulated as $\textbf x^{L+1} = f_k(\textbf x^L) \otimes f_q(\textbf x^L) \otimes f_v(\textbf x^L)$, which is also element-wise multiplication between processed input feature themselves. 
Characterized by element-wise interaction, these designs offer complementary feature correlation capabilities in addition to simple linear superposition.
Such feature interactions have become the essential mechanisms of mainstream state-of-the-art neural architectures.
For example, it's implemented in SENet with squeeze and excitation~\cite{hu2018squeeze}, non-local network with transposed multiplication~\cite{wang2018non}, vision transformers with self-attention~\cite{vaswani2017attention,bai2021attentions,dosovitskiy2020image}, gated aggregation~\cite{li2022efficient,rao2022hornet,yang2022focal}, and quadratic neurons~\cite{xu2023quadranet,fan2023one,xu2022quadralib}. 


Although these models greatly improve the performance of state-of-the-arts, as mentioned above, these works provide diversified explanations that neglect the underlying shared design of element-wise feature multiplication operation~\cite{ma2024rewrite}, and thus may fail to reveal the fundamental source of improvement.
To provide both explainability and quantifiability, in this paper, we propose a unified theoretical perspective to rethink the feature interaction scaling, i.e., \textit{the dimensionality of feature interaction space}.
For example, as shown in Fig~\ref{fig:1}(b), by employing the $\otimes$, element-wise multiplication, an implicit interactive quadratic space $\mathcal{Q} = \text{span}\left(x_1^2, x_1x_2,\cdots, x_{n-1}x_n, x_n^2\right)$ with degrees of freedom ${n(n+1)}/{2}$ is constructed from the original representative vector space $\mathcal{V} =\text{span}\left(x_1,x_2,\cdots,x_n\right)$ with degrees of freedom $n$~\cite{ma2024rewrite}.
Such space dimensionality scaling is the key to improving feature representation quality and end-task, as we will show later. 

From the unified feature dimensionality perspective, a new opportunity emerges in neural architecture design, that is to scale to \textit{infinite-dimensional feature interactions} than former methodologies (e.g., from $n(n+1)/2$ to $\lim_{k\to \infty} \frac{(n+k-1)!}{(n-1)!k!}$). 
However, scaling feature interaction space dimensionality from architectural enhancements (e.g., quadratic, self-attention, and recursive gates) comes with linear scaling cost w.r.t. interaction order $k$, which hinders the infinite dimensionality increase~\cite{li2022efficient,dosovitskiy2020image}. Thus, there is an open question:
\begin{center}
\vspace{-1mm}
\textit{How can we efficiently extend interactions to an infinite-dimensional space?}
\end{center}
\vspace{-1mm}
Inspired by traditional machine learning, we propose an approach that introduces kernel methods for feature interaction in neural networks. We define a set of feature interactions between features $W_a\mathbf{x}$ and $W_b\mathbf x$ with a kernel function $\mathit K(W_a\mathbf{x},W_b\mathbf x)$ instead of the element-wise multiplication.
As shown in Fig~\ref{fig:1}(c), the kernel method transforms the feature to an ultra-high dimensional space by an implicit mapping $\phi(\cdot)$.
From there, the feature interaction space is defined by the inner product on the Reproducing Kernel Hilbert Space (RKHS)~\cite{aronszajn1950theory} $\mathcal H$ constituent with the kernel function $\mathit K(\cdot,\cdot)$. The RKHS can greatly expand the interaction space at very little cost, enabling infinite dimensions.

We then propose InfiNet, a novel family of neural networks that generate high-quality feature representations. 
    Specifically, we introduce the Radial Basis Function (RBF) kernel~\cite{musavi1992training} as a replacement of the common $\oplus$ or $\otimes$ operations. With the infinite series expansion in RBF kernel, it enables a theoretical provable dimensionality approximation $\text{span}_{j \in \mathbb N,\sum_k{n_k}=j}\{\frac{x_1^{n_1} \cdots x_k^{n_k}}{\sqrt{n_{1}!\cdots n_{k}!}}\}$ while with as low-overhead as evaluating an exponential function. 
    In this way, InfiNet enables efficient infinite-dimensional feature interaction space scaling upon a finite set of branches in the model architecture, thus achieving better complexity performance tradeoffs than prior state-of-the-art.

\textbf{Contributions.} We make the following contributions: 
\vspace{-1mm}
\begin{itemize}
    \item We unify the perspectives of recent feature interactive works and identify a novel direction of neural network performance scaling: the feature interaction space dimensionality. 
    \item We propose a method to expand the feature interaction space to an infinite dimension with RBF kernel, that can effectively model the complex implicit correlations of features.
    \item We propose InfiNet, a novel series of neural networks that explore the neural interaction from infinite-dimensional space, and achieve state-of-the-art performance.
\end{itemize}

\vspace{-1mm}
Extensive ablation studies verify that the shift from finite feature interaction space to infinite one is a key factor to learn better representations and therefore improving model performance.
Meanwhile, large-scale experiments on ImageNet classification, MS COCO detection and ADE20K segmentation also demonstrate InfiNet design's effectiveness, which consistently outperforms the state-of-the-art flat stream networks~\cite{liu2022convnet,he2016deep} and finite-order interaction networks~\cite{rao2022hornet,liu2021swin}.

%% file: _txt/2_Related_Work.tex
\section{Related Work}

\textbf{Interactions in Neural Networks.} Interaction in neural networks has undergone a long period of change. 
During the evolution from AlexNet~\cite{krizhevsky2012imagenet} to ResNet~\cite{he2016deep}, the model has for long followed the principle of summing the weighted pixels at each position in the layer-by-layer feature iteration. And then, as the potential of the attention mechanism was realized, model development is towards an element-wise multiplication way. This trend is punctuated by the emergence of models such as Non-Local~\cite{wang2018non}, Transformer~\cite{vaswani2017attention}, and ViT~\cite{dosovitskiy2020image}.
Recent insights suggest that the crux of the attention mechanism lies in high-order information interactions, rather than the mechanism of "Attention" itself~\cite{bai2021attentions,rao2022hornet,xu2024quadranet}. 
This revelation has spurred the development of innovative neural network architectures. For example, HorNet~\cite{rao2022hornet}, QuadraNet~\cite{xu2023quadranet} and MogaNet~\cite{li2022efficient} examines high-order models from the perspective of spatial interactions through multiplication-integrated architecture design. However, limited by the support of existing deep learning platforms such as Pytorch~\cite{paszke2019pytorch}, few attempts have been made to extend the model's feature interaction to an ultra-high dimensional situation through the kernel method for more potential.

\textbf{Kernel Methods in Neural Networks.} 
The fundamentals of kernel methods are well-studied in traditional ML domains like support vector machines \cite{scholkopf1998} but they are less used in neural architectures. Early extensions to deep learning through the kernelized perceptron \cite{cho2009} have improved the performance but mainly in shallow neural networks. 
Recent advancements include the development of Convolutional Kernel Networks (CKN) \cite{mairal2014}, which merge CNNs' robust feature learning with kernel stability. This approach offers a theoretical foundation for deep learning's application in structured data. 
Additionally, the introduction of Kervolutional Neural Networks \cite{chen2019} replaces traditional convolution in CNNs with kernel-based operations to enhance feature extraction without excessive computational costs. 
The combination of Gaussian processes with neural networks to create Deep Kernel Learning \cite{wilson2016} adjusts model complexity based on data while maintaining Bayesian inference. The neural tangent kernel (NTK)~\cite{jacot2018neural} framework establishes a direct connection between infinitely wide neural networks at initialization and kernel methods. Specifically, NTK shows that as the width grows, the network's training dynamics can be described by a kernel function, linking the neural network's behavior to that of kernel methods. Random features~\cite{rahimi2007random} provide an efficient approximation to the feature mappings used in kernel methods. By random projections, one can approximate the inner product defined by a kernel function, making it feasible to apply kernel methods~\cite{hashemi2023generalization,bach2008exploring}. 

Despite these innovations, a significant limitation remains: Although these methods expand the feature representation space, they fail to scale up the feature interaction space, limiting the network to aggregate information in a superposition manner. 
As a result, these methods also fall short of potential feature interactions and thus are incapable of handling complex data correlations and functionalities.


%% file: _txt/3_SpaceTransfer.tex
\section{From Feature Representation Space to Interaction Space}
\label{sec:3}
We start with considering the transformation of feature representation of a normal shape-preserving 2-layer perceptron block. Given an input $\mathbf x = (x_1, x_2, \cdots, x_n)$ with n-dimension. We denote the first layer transformation as $\mathbf z =g(\mathbf x)=\sigma(W_1\mathbf x)$ (we omit the bias term for simplicity), and the second layer transformation as $h(\mathbf z)=W_2\mathbf z$. Therefore the whole block is a mapping $h\circ g: \mathbb R^n \mapsto \mathbb R^n \mapsto \mathbb R^n$ from feature space $\mathbb R^n$ to a middle feature representation space $\mathbb R^n$ and eventually to a output space $\mathbb R^n$. Expanding the width of the neural networks, for example with a coefficient 2, is going to expand the feature representation space to $\mathbb R^{2n}$. The core idea of this dimensional expansion is that implicit associations in features in low dimensions will be expressed explicitly when projected into a high-dimensional space. Model architecture (i.e. convolution, multi-branch~\cite{szegedy2015going}) is an additive superposition of pixels, essentially no different from the perceptron in terms of representation space.
\subsection{Feature Interaction Space}
Although flat stream neural networks are defined by linear transformations and activations, modern network design also introduces multiplications in the network structure or neurons. They are called attentions from an interpretable point of view, that is the degree of interest of one position in relation to another; from a more abstract point of view, they are called interactions in spatial contexts. But invariably, these are realized with element-wise multiplication. We conduct the following definition.

\textbf{Definition 1} (Feature Interaction) A Feature interaction refers to transformations between features with the same or different positions defined by element-wise multiplication.

For example, Star Operation~\cite{ma2024rewrite} is a basic 2-order feature interaction, defined as $(W_a\mathbf x)*(W_b\mathbf x)$, where $W_a,W_b \in \mathbb R^n$. This is a simple element-wise multiplication fusion of two linear transformations of the ordinary input $\mathbf x$.  We write the expansion of such multiplication operation: 
\begin{equation}
     y =  (W_a\mathbf x)*(W_b\mathbf x) = (\sum_{i=1}^{n} w_{ai}x_i)(\sum_{j=1}^{n} w_{bj}x_j) = \sum_{i \leq j}  \alpha_{i,j} x_ix_j 
     \label{equ:ewm}
\end{equation}
where $\alpha_{i,j} = w_{ai}w_{bj}+w_{aj}w_{bi},\text{if}\  i\neq j$, and $\alpha_{i,i} = w_{ai}w_{bi}$. Then we vectorize $\alpha$ and $x_ix_j$:
\begin{equation}
    A=[\alpha_{1,1},\alpha_{1,2},\alpha_{2,2},\cdots,\alpha_{n-1,n},\alpha_{n,n}] \in \mathbb R^{n(n+1)/2}
\end{equation}
\begin{equation}
    \chi = [x_1x_1,x_1x_2,x_2x_2,\cdots,x_{n-1}x_n,x_nx_n]
\end{equation}
$\chi$ can therefore define a basis of a space. Thus the output of the current layer can be rewritten as:
\begin{equation}
    y = \sum_{i \leq j}  \alpha_{i,j} x_ix_j = A\chi. 
    \label{equ:y}
   \end{equation}
From the independence of the pixel level, we know that each term in $\chi$ is linearly independent, this indicates every dimension in $\chi$ is an individual dimension. Given a set of basis vectors $\chi$, we define $\text{span}(\chi)$ as the feature interaction space. 
In this way, the generation of the next layer of features $y$ is constituted by a linear superposition $A\chi$ on the feature interaction space $\text{span}(\chi)$ like in Eq.(\ref{equ:y}).

\subsection{Dimension of Feature Interaction Space}
Now, we consider the number of dimensions of a feature interaction space. Given $k$-1 multiplication operations, we first define the \textit{k}-order feature interaction space as follows:

\textbf{Definition 2} (Feature Interaction Space) A \textit{k}-order Feature Interaction Space $\mathcal S^k$ refers to the span of monomial basis $\{x_1^{d_1}x_2^{d_2}\cdots x_n^{d_n}| \sum d_i = k, d \in \mathbb N\}$ defined by a \textit{k}-order Feature Interaction. 

An element-wise multiplication generates a feature interaction space of $n(n+1)/2$ dimension, which is the number of elements of an upper triangular matrix. In general, considering the symmetry of the interactions and elements, for a k-order interaction on the $n$-dimensional feature, the dimension of the corresponding feature interaction space is:
\begin{equation}
    dim(\mathcal S^k) = \frac{(n+k-1)!}{(n-1)!k!}
    \label{equ:dim}
\end{equation}
The next layer of feature generation based on the feature interaction space greatly expands the spatial dimensions to which the features are mapped compared to the original model based only on the feature representation space, i.e., it is possible to explore the feature's non-linearity in high-dimensional space. It is worth noting that this process introduces terms like $x_1x_2$, an interaction that cannot be captured by traditional plane networks in feature representation space.

For example, we consider the Self-Attention in the Transformers~\cite{vaswani2017attention}. The Self-Attention contains two element-wise multiplications, so it explores a 3-order feature interaction space. This is due to the fact that: (1) in the first stage, in the query-key dot-product attention map computation $Att(\mathbf x) = Q\cdot K^T = W_Q\mathbf x\cdot\mathbf x^TW_K^T$ explores the feature interaction space $\mathcal{A} = \text{span}\left(x_1^2, x_1x_2,\cdots, x_{n-1}x_n, x_n^2\right)$. (2) In the second stage, the multiplication between the attention map and the value $y = \sigma(Att(\textbf x))\cdot V = \sigma(Att(\textbf{x}))\cdot W_v\mathbf x$
explores the feature interaction space $\mathcal S = \mathcal A \otimes \mathbb R^n = \text{span}\left(x_1^3, x_1^2x_2,x_1x_2x_3,\cdots, x_{n-1}x_n^2, x_n^3\right)$, which has $(n+2)(n+1)n/6$ dimensions in line with Eq.(\ref{equ:dim}). Thereby we explain, from the perspective of feature interaction space, why the transformer family of models has, so far, generally outperformed recurrent neural networks and convolutional neural networks in various domains.

\section{Expanding Interaction Space to Infinite Dimension}

\label{sec:4}
This element-wise multiplication-based interaction, since the construction of each order of the feature interaction space is based on a multiplication operator, leads to a problem in that feature interaction space expansion is still difficult. This is due to the fact that these interaction operators is explicit mapping, which tends to have quadratic or higher complexity w.r.t the input length (e.g. self-attention~\cite{vaswani2017attention}) or lengthy recursive designs (e.g. HorNet~\cite{rao2022hornet}) and linear complexity w.r.t interaction order.
But the problem with building this mapping explicitly is: (1) The complexity of mapping itself. The computational overhead associated with defining a set of explicit nonlinear mappings is non-negligible. A mapping from $C$ channels to $C'$ channels means a computation complexity of $O(CC')$. (2) The complexity of interaction. The complexity of the inner product used to interact with the two sets of features increases dramatically to $O(C')$ when the dimension is raised. Considering $C'>>C$, these two complexities will largely increase the computational overhead of the networks.

We would like to obtain a method that can expand the dimension of feature interaction space as much as possible in $O(1)$ time. Fortunately, the machine learning community has already given a method for increasing the dimension of a feature defined on the inner product: \textbf{kernel methods}. 

\subsection{Expanding Interaction Space with Reproducing Kernel}

The nature of kernel methods is that they are substitutions for inner product operations. This requires combining element-wise multiplication and summation to define a set of inner products within the network.
For this purpose,
we rewrite the form of element-wise multiplication in Eq.(\ref{equ:ewm}) on two groups of features, which is a common design in literature architecture (e.g. multi-head self-attention~\cite{vaswani2017attention}).
It will therefore be a superposition of multiple interaction in the format:
\begin{equation}
    y = \sum_{i=1}^{C}W_{ai}\mathbf x * W_{bi}\mathbf x = \langle \mathbf{W_ax,W_bx}\rangle
\end{equation}
where $\mathbf{W_ax} = [W_{a1}\mathbf{x}, W_{a2}\mathbf{x}, \cdots, W_{aC}\mathbf{x}]$, $C \in \mathbb N$ is the number of branches. Thereby we generalize the form of the feature interaction from element-wise multiplication to inner product which is a multi-branch paradigm. At this point, we have $\mathbf{W_ax} \in \mathbb R^C$, while it is generated from a feature representation space $\mathbb R^n$. In order to extend the interaction space, we need to further project $\mathbf{W_ax}$ and $\mathbf{W_bx}$ to a high-dimensional space. An obvious way to do this is to construct an implicit mapping $\Phi(\cdot)$ to a high-dimensional space, so we can compute $\langle \Phi(W_a\mathbf x),\Phi(W_b\mathbf x)\rangle$ for interaction.

By Mercer's Theorm~\cite{mercer1909xvi}, taking a continuous symmetric positive semi-definite function $K(s,t)$, there is an orthonormal basis $\{\phi_i(\cdot)\}$, $i =0,1,\cdots,\infty$, consisting of eigenfunctions of function $K(\cdot,\cdot)$ such that the corresponding sequence of eigenvalues $\{\lambda_i\}$ is non-negative. These means:
\begin{equation}
    K(s,t) = \sum_{i=1}^{\infty}\lambda_i\phi_i(s)\phi_i(t),
\end{equation}
where $\forall i\neq j, \forall s \ \text{and}\ t, \langle\phi_i(s),\phi_j(t)\rangle = 0$ since. Then we construct a Hilbert space $\mathcal H$ with the orthonarmal basis $\{\sqrt{\lambda_i}\phi_i(\cdot)\}$. Consider a vector $f = (f_1,f_2,\cdots)^T_{\mathcal H}$ on the space $\mathcal H$, then we have: 
\begin{equation}
    f = \sum_{i=1}^{\infty}f_i\lambda_i\phi_i(\cdot).
\end{equation}
Thus for a vector $K(s,\cdot)$ in space $\mathcal H$:
\begin{equation}
    K(s,\cdot) = \sum_{i=1}^{\infty}\lambda_i\phi_i(s)\phi_i(\cdot)=\sum_{i=1}^{\infty}\sqrt{\lambda_i}\phi_i(s)\sqrt{\lambda_i}\phi_i(\cdot)=(\sqrt{\lambda_1}\phi_1(s),\sqrt{\lambda_2}\phi_2(s),\cdots)^T_{\mathcal H}.
\end{equation}
Therefore, for the Hilbert space $\mathcal H$, we can define the reprodcuing kernel by:
\begin{equation}
    \langle K(s,\cdot),K(t,\cdot) \rangle = \sum_{i=1}^{\infty}\sqrt{\lambda_i}\phi_i(s)\sqrt{\lambda_i}\phi_i(t) =  \sum_{i=1}^{\infty}\lambda_i\phi_i(s)\phi_i(t).
\end{equation}
\textbf{Implicit Mapping to High-Dimensional RKHS}. Let $\Phi(s) =  K(s,\cdot)$, then we have $ \langle \Phi(s),\Phi(t) \rangle = K(s,t)$. The Hilbert Space $\mathcal H$ is known as the Reproducing Kernel Hilbert Space (RKHS) corresponding to kernel function $K(\cdot,\cdot)$. Note that the $\Phi(\cdot)$ is therefore defined on the RKHS $\mathcal H$, which can be infinite-dimensional given specific kernel $K(\cdot,\cdot)$. The mapping $\Phi(\cdot)$ does not have to have an explicit expression since we can get the result of $\langle \Phi(s),\Phi(t) \rangle$ by computing $K(s,t)$. This means that we can achieve an extension of the dimensions for the feature interaction space by simply replacing the inner product $\langle \mathbf{W_ax,W_bx} \rangle$ used in the interaction with a kernel $K(\mathbf{W_ax,W_bx})$.

\subsection{Infinite-Dimensional Feature Interaction with RBF Kernel}

To maximize the dimension of the feature interaction space, we consider Radial Basis Function (RBF) Kernel $K_{\text{rbf}}(\mathbf s,\mathbf t) =\exp  \left(-\frac{1}{2}\left\|\mathbf{s}-\mathbf{t}\right\|_2^2\right)$with an infinite-dimensional RKHS, given the fact that: 
\begin{align}
\exp & \left(-\frac{1}{2}\left\|\mathbf{s}-\mathbf{t}\right\|_2^2\right)=\sum_{j=0}^{\infty} \frac{\left(\mathbf{s}^{\top} \mathbf{t}\right)^j}{j !} \exp \left(-\frac{1}{2}\|\mathbf{s}\|_2^2+\left\|\mathbf{t}\right\|_2^2\right)\\
=& \sum_{j=0}^{\infty} \sum_{n_1+n_2+\cdots+n_k=j} \exp \left(-\frac{1}{2}\|\mathbf{s}\|^2\right) \frac{s_1^{n_1} \cdots s_k^{n_k}}{\sqrt{n_{1}!\cdots n_{k}!}} \exp \left(-\frac{1}{2}\left\|\mathbf{t}\right\|^2\right) \frac{t_1^{ n_1} \cdots t_k^{n_k}}{\sqrt{n_{1}!\cdots n_{k}!}}
\\
=& \langle \Phi_{\text{rbf}}(\mathbf s), \Phi_{\text{rbf}}(\mathbf t) \rangle,
\end{align}
where $\Phi_{\text{rbf}}(\mathbf x) = \sum_{j=0}^{\infty} \sum_{\sum_kn_k=j} \exp \left(-\frac{1}{2}\|\mathbf{x}\|^2\right) \frac{s_1^{n_1} \cdots s_k^{n_k}}{\sqrt{n_{1}!\cdots n_{k}!}} \in \text{span}_{j \in \mathbb N,\sum_k{n_k}=j}\{\frac{x_1^{n_1} \cdots x_k^{n_k}}{\sqrt{n_{1}!\cdots n_{k}!}}\}$. 

\textbf{Infinite-dimensional Feature Interaction Space} Observing the RKHS of such a RBF kernel, $\text{span}_{j \in \mathbb N,\sum_k{n_k}=j}\{\frac{x_1^{n_1} \cdots x_k^{n_k}}{\sqrt{n_{1}!\cdots n_{k}!}}\}$, We note that each of its dimensions is a j-order interaction within the feature $\mathbf x$, given the fact one of the bases of this RKHS is:
\begin{equation}
    \{1, x_1,\cdots ,x_n,x_1^2,\cdots,x_1x_n,\cdots,x_n^2,\cdots,x_1^j,x_1^{j-1}x_2,\cdots,x_n^j,\cdots\},
\end{equation}
which contains an all-order monomial among all elements of the feature $\mathbf x$. This means that we get an infinite-dimensional Hilbert space for the superposition of interaction information through such an RBF kernel, and most importantly, each dimension of this space is defined by a feature interaction. Therefore, we get an infinite-dimensional feature interaction space.

\begin{figure}[t]
    \centering
    \includegraphics{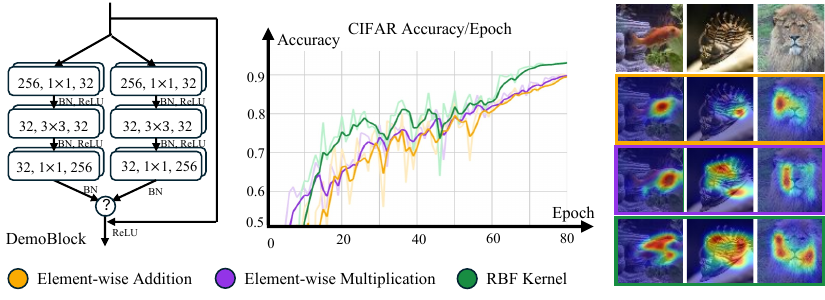}
    \caption{Comparison of simple representation, finite interaction, and infinite-dimensional interaction. The \textbf{?} circle in DemoBlock is chosen from element-wise Add, element-wise Mul. or RBF kernel. }
    \label{fig:2}
\end{figure}

\subsection{Demo Case Performance of Models on Different Feature Space}
In order to compare networks that utilize the summing superposition of information mappings on the feature representation space, finite feature interaction space, and infinite-dimensional interaction space, we design a demo network with 8 DemoBlock, as shown in Fig.~\ref{fig:2}. 
DemoBlock has a two-group design, the only difference among models in different spaces is the interaction method at the end of the block (add for simple representation, multiplication for finite interaction, and RBF kernel for infinite dimensional interaction), the demo models are trained on CIFAR10 and Tiny-ImageNet.

As shown in Fig.~\ref{fig:2}, with the procedure of transferring from the simple feature representation space to a finite feature interaction and eventually to an infinite-dimensional interaction, the performance on CIFAR10 of the networks is growing. This is done throughout the training process, implying the superiority of feature iteration in a high-dimensional interaction space. The right side of Fig.~\ref{fig:2} shows the Class Activation Mapping \cite{zhou2016learning} of three different demo nets on Tiny-ImageNet. From this, we can see that the network of feature interaction better reflects the pixel-level correlation within the image.



%% file: _txt/4_InfiNet.tex
\section{Method}
\begin{figure}[t]
    \centering
    \includegraphics{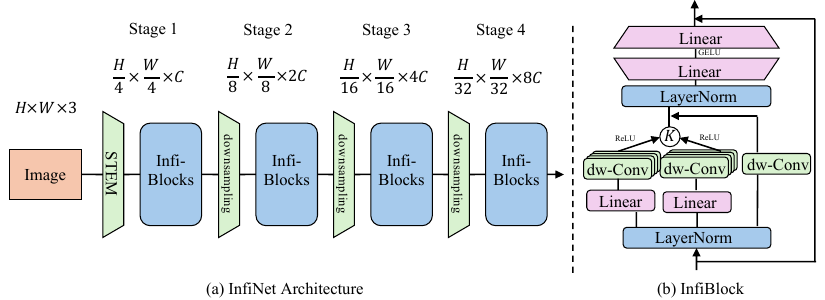}
    \caption{\textbf{Overview of InfiNet.} (a) Four-stage hierarchical InfiNet design. (b) InfiBlock Design}
    \label{fig:3}
\end{figure} 

\subsection{InfiBlock: Infinite-Dimensional Spatial Feature Interaction}

\textbf{InfiBlock.} In this section, we present InfiBlock, the basic block to build the high-performant InfiNet architecture to achieve infinite-dimensional spatial feature interaction. As presented in Fig.~\ref{fig:3}(b), InfiBlock starts with a LayerNorm layer and subsequently transforms the feature into separate representations of the two groups through two different linear layers. Then InfiBlock utilizes a depth-width convolution with an expansion coefficient of $r$ and a ReLU activation to obtain a feature branch vector of length $r$ on each group. The feature branch vectors on both groups are then fed into an RBF kernel for feature interaction. At the same time, we retain a pathway containing only one depth-wise convolution for summing superposition over the feature representation space to ensure that the linear connections between features are not neglected and overfitted. This is followed by a residual connection with the original input values after passing through a two-layer MLP. Starting with a input $X^l \in \mathbb R^{HWC}$, Infi-Block can be formulated as:
\begin{align}
   & \hat X^l = \text{LayerNorm}(X^l),\\
   & [\vec Z_a^l,\vec Z_b^l, Z_c^l]= [\sigma(\text{Conv}_a(\hat X^lW_a)), \sigma(\text{Conv}_b(\hat X^lW_b)), \text{Conv}_c(\hat X^l)],\\
   & X^{l+1} = \text{MLP}(\text{LayerNorm}(K_{rbf}(\vec Z^l_{a},\vec Z^l_{b})+Z_c^l)) + X^l,
\end{align}
where $K_{rbf}(\vec Z^l_{a},\vec Z^l_{b}) = exp(-\frac{\|\vec Z_a^l-\vec Z_b^l\|_2^2}{2}) \in \mathbb{R}^{HWC}$, is an RBF kernel with out any hyper-parameter, and $\|\cdot\|_2^2$ is squared L-2 norm. $\text{Conv}_{a,b}(\hat X^lW_{a,b}) \in \mathbb{R}^{HWC*r} $ expands the number of channels of feature to r times, where $W_{a,b}$ preserve the shape of the feature. The discretization of the feature is performed in a depth-width convolution. $\text{LayerNorm}(\cdot)$ is layer normalization~\cite{ba2016layer} and $\text{MLP}(\cdot)$ is activated by GELU~\cite{hendrycks2016gaussian}.

Specifically, following the ConvNeXt~\cite{liu2022convnet}, we select a $7\times7$ depth-wise convolution to obtain a larger receptive field. We set the expansion coefficient $r$ as 7 in practice. This means: 
\begin{equation}
       [\vec Z_a^l,\vec Z_b^l]= [[Z_{a1}^l,Z_{a2}^l,\cdots,Z_{a7}^l]^T,[Z_{b1}^l,Z_{b2}^l,\cdots,Z_{b7}^l]^T],
\end{equation}
where $ Z_{ai}^l = \sigma(\text{DW-Conv}_{ai}(X^lW_a))\ \in \mathbb R^{HWC}$. Therefore, the RBF kernel performs feature interaction on multiple (7) branches of convolution filter and outputs a result still in $\mathbb R^{HWC}$.

\subsection{Model Architectures}

As shown in Fig.~\ref{fig:3}(a), InfiNet uses the widely adopted 4-stage hierarchical architecture as in ResNet~\cite{he2016deep}, ConvNeXt~\cite{liu2022convnet} and Swin Transformer~\cite{liu2021swin}. We stack InfiBlocks into each stage. We set the number of channels in each stage as [C,2C,4C,8C] following common practice. We build a family of InfiNets that InfiNet-T/S/B/L/XL with model variants hyper-parameters: $C=\{64,96,128,128,192\}$, number of blocks in each stages  $=\{2,2,18,2\}$ for InfiNet-T/S/B and $\{3,3,27,3\}$ for InfiNet-L/XL. A down-sampling with $2\times2$ convolution with stride = 2 is used to connect each stage. STEM~\cite{liu2022convnet} is used to connect the input of InfiNet to Stage 1.

%% file: _txt/5_Experiments.tex
\section{Experiments}
\label{sec:6}

We perform a series of experiments to validate the efficacy of InfiNets. The primary results on ImageNet~\cite{deng2009imagenet} are showcased and contrasted with several architectures. Furthermore, our models were evaluated on downstream ADE20K\cite{zhou2017scene} semantic segmentation and COCO~\cite{lin2014microsoft} object detection. ImageNet-1K experiments are conducted on 4$\times$Nvidia A100 GPUs and ImageNet-21K on 16$\times$.

\vspace{-2mm}
\subsection{ImageNet Classification}

\textbf{Setups.} We conduct image classification experiments on ImageNet-1K~\cite{deng2009imagenet}, which contains 1.28 million training samples belonging to 1000 classes and 50K samples for validation. We train the InfiNet-T/S/B/L models for 300 epochs with AdamW~\cite{loshchilov2018decoupled} optimizer. We use the cosine learning rate scheduler~\cite{loshchilov2017sgdr} with 20 warmup epochs and the basic learning rate is set as $4\times10^{-3}$. The training resolution is set as $224\times224$. To further evaluate the InfiNet's scalability, we train the InfiNet-L/XL models on 14M-sample ImageNet-22K dataset for 90 epochs and then fine-tune on ImageNet-1K at $384\times384$ resolution for 30 epochs following~\cite{liu2022convnet}. More details can be found in Appendix~\ref{app:training}.

\textbf{Results.} We present our ImageNet experiment results and comparison with baselines in Table~\ref{tab:result}. Our models achieve competitive performance with state-of-the-art. It is worth noting that InfiNet has about 20\% fewer FLOPs\footnote{Difference between FLOPs and FLOPS: FLOPs (floating point of operations), the number of floating point operations, is used to measure the complexity of an algorithm/model. This number gets smaller the better. The FLOPs is different from FLOPS (floating point per second), which is a measure of hardware performance. Following other deep-learning architecture works, we use FLOPs to measure the computing demands. Since our model gets a smaller FLOPs, our method is more efficient.} than the baseline models with similar parameter scales, but still achieves great performance. It demonstrates the effectiveness of our proposed generation of features in infinite-dimensional interaction spaces. Experiments on isotropic models can be found in Table~\ref{tab:ablation}(a).

\input{_tab/ImageNet}

\vspace{-2mm}
\subsection{MS COCO Detection}
\textbf{Setups.} We evaluate our models for object detection tasks on widely used MS COCO~\cite{lin2014microsoft} benchmark. In the experiments, InfiNet-T/S/b/XL serves as the backbone network within Cascade Mask RCNN~\cite{cai2018cascade}. We use AdamW as the optimizer and a batch size of 16 and adhere to the 3$\times$ schedule, following ConvNeXt~\cite{liu2022convnet} and Swin~\cite{liu2021swin}. We resize the input so that the longer side is at most 1333 and the short side is at most 800. We initialize the backbone model with ImageNet-1K pre-trained weights for T/S/B models and ImageNet-22K pre-trained weights for XL model. 

\textbf{Results.} As shown in Table~\ref{tab:detection}, InfiNets comprehensively beat the non-interactive model ConvNeXt~\cite{liu2022convnet}, and space-limited interactive Swin~\cite{liu2021swin} and HorNet ~\cite{rao2022hornet} under the same cascade Mask RCNN framework in box AP and mask AP. This means that for such dense prediction tasks, spatial interaction of features in a high-dimension space is crucial. The InfiNet series model obtain$ 0.9\sim1.5$ box AP and $1.3\sim2.5$ mask AP gain compared with non-interactive ConNeXt.

\input{_tab/COCODetection}
\input{_tab/Ablation}

\subsection{ADE20 Segmentation}

\textbf{Setups.} We evaluate our models for the semantic segmentation task on widely used ADE20K~\cite{zhou2017scene} benchmark covering 150 semantic categories on 25K images, in which 20K are used for training. We use UperNet~\cite{xiao2018unified} for as the basic framework and adopt InfiNets as the backbone model. Training details follow the Swin~\cite{liu2021swin}, we use AdamW optimizer with learning rate $1\times10^{-4}$ and batch size 16. 

\textbf{Results.} The right half of Table~\ref{tab:detection} lists the mIoU and corresponding model size and FLOPs for different configurations. Our models beat most of the baseline in the segmentation task. The results show that as the model size increases, the performance gap between InfiNet and other baselines is getting larger, illustrating the scalability of InfiNet on segmentation.

\subsection{Ablation Study}
We use the additive operator, Hadamard product, quadratic polynomial kernel and cubic polynomial kernel, and RBF kernel in the kernel methods section of InfiNet, respectively, on a tiny size model, to verify the effect of the gradual expansion of the order of the interaction space up to infinite dimensions on the performance of the model. As in Table~\ref{tab:ablation}(b), we can see that the performance of the model is gradually improving as the order of the model interaction space increases up to infinite dimensions.

%% file: _tab/ImageNet.tex
\begin{table}
  \caption{\textbf{ImageNet classification results.} We compare our models with state-of-the-art models with comparable parameters, the Top-1 accuracy is reported on the ImageNet-1K validation set.}
  \label{tab:result}

  \begin{subtable}{.5\linewidth}
  \centering
  \tabcolsep 2pt
  \resizebox{!}{.58\linewidth}{
  \begin{tabular}{lllll}
    \toprule
        &  Interact.  & Params& FLOPs& Top1 \\
      Model     & Orders  & (M)  & (G) & Acc.(\%) \\
    \midrule
        ConvNeXt-T\cite{liu2022convnet} & no& 29 & 4.5 & 82.1\\
         SLaK-T\cite{liu2022more} & no & 30 & 5.0 & 82.5\\
        Conv2Former-T\cite{hou2024conv2former} &2 & 27 &4.4 & 83.1\\
         UniFormer-S\cite{li2021uniformer} &2 & 22 & 3.6 & 82.9 \\
                 CoAtNet-0\cite{dai2021coatnet} & 3 & 25&4.2 &82.7\\
        FocalNet-T\cite{yang2022focal}&3& 28 &4.4 &82.1\\
                Swin-T\cite{liu2021swin} &3& 28&4.5&81.3\\
        HorNet-T\cite{rao2022hornet}&2-5& 22 & 4  &82.8  \\
        MogaNet-S\cite{li2022efficient}&4 & 25 & 5.0  &83.5  \\

  \rowcolor{mygray} \textbf{InfiNet-T} &$\infty$  & 23  & 3.2& 83.4  \\
  \midrule
          ConvNeXt-S\cite{liu2022convnet} & no& 50 & 8.7 & 83.1\\
         SLaK-S\cite{liu2022more} & no & 55 & 9.8 & 83.8\\
        Conv2Former-S\cite{hou2024conv2former} &2 & 50 & 8.7 & 84.1\\
         UniFormer-B\cite{li2021uniformer} &2 & 50 & 8.3 & 83.9 \\
                 CoAtNet-1\cite{dai2021coatnet} & 3 & 42&8.4 &83.3\\
        FocalNet-S\cite{yang2022focal}&3& 50 &8.7 &83.5\\
                Swin-S\cite{liu2021swin} &3& 50& 8.7&83.0\\
        HorNet-S\cite{rao2022hornet}&2-5& 50 & 8.8  &84.0  \\
        MogaNet-B\cite{li2022efficient}&4 & 44 & 9.9  &84.3  \\
   \rowcolor{mygray} \textbf{InfiNet-S} &$\infty$  & 48  & 7.2 & 84.0 \\

    \bottomrule
  \end{tabular}}
  \end{subtable}
    \begin{subtable}{.5\linewidth}
  \centering
    \tabcolsep 2pt
    \resizebox{!}{.58\linewidth}{
  \begin{tabular}{lllll}
    \toprule
     &  Interact.  & Params& FLOPs& Top1 \\
   Model        & Orders  & (M)  & (G) & Acc.(\%) \\
    \midrule
     ConvNeXt-B\cite{liu2022convnet} & no& 89 & 15.4 & 83.8\\
         SLaK-B\cite{liu2022more} & no & 85 & 17.1 & 84.0\\
        Conv2Former-B\cite{hou2024conv2former} &2 & 90 & 15.9 & 84.4\\
                 CoAtNet-2\cite{dai2021coatnet} & 3 & 75&15.7&84.1\\
        FocalNet-B\cite{yang2022focal}&3& 89 &15.4 &83.9\\
                Swin-B\cite{liu2021swin} &3& 89 &15.4&83.5\\
        HorNet-B\cite{rao2022hornet}&2-5& 87 & 15.6  &84.3  \\
        MogaNet-L\cite{li2022efficient}&4 & 83 & 15.9  &83.5  \\
    \rowcolor{mygray} \textbf{InfiNet-B}    &$\infty$  & 82   & 12.8 & 84.5 \\
     \rowcolor{mygray} \textbf{InfiNet-L}&$\infty$ & 116.8   & 19.1 & 84.8 \\
    \midrule
   \multicolumn{4}{l}{\small \textit{ImageNet-21K Pretrained Models Fine-tuned @$384^2$}}\\
    \midrule
                  ConvNeXt-L$^\ddagger$\cite{liu2022convnet} & no& 198 & 101 & 87.5\\
                 CoAtNet-3$^\ddagger$\cite{dai2021coatnet} & 3 & 168&107 &87.6\\
        FocalNet-L$^\ddagger$\cite{yang2022focal}&3& 197 &101 &87.3\\
                Swin-L$^\ddagger$\cite{liu2021swin} &3& 197& 104&87.3\\
        HorNet-L$^\ddagger$\cite{rao2022hornet}&2-5& 202 & 102  &87.7  \\
        MogaNet-XL$^\ddagger$\cite{li2022efficient}&4 & 181 & 102  &87.8  \\
       \rowcolor{mygray} \textbf{InfiNet-L$^\ddagger$}  &$\infty$  & 116.8   & 60 & 87.8 \\
        ConvNeXt-XL$^\ddagger$\cite{liu2022convnet} & no& 350 & 179 & 87.8\\
     \rowcolor{mygray} \textbf{InfiNet-XL$^\ddagger$}   &$\infty$   & 255.8   & 126 & 88.2 \\
  
    \bottomrule
  \end{tabular}}
  \end{subtable}
\vspace{-3mm}
\end{table}

%% file: _tab/COCODetection.tex
\begin{table}
  \caption{\textbf{Object detection and semantic segmentation results on MS COCO and ADE20K.}}
  \label{tab:detection}
    \resizebox{\textwidth}{!}{
  \centering
  \begin{tabular}{lcccccccc}
    \toprule
&\multicolumn{4}{c}{Object Detection with \textit{Cascade Mask R-CNN 3×}}&\multicolumn{4}{c}{Semantic Segmentation with \textit{UperNet 160K}}\\
\cmidrule(lr){2-5}\cmidrule(lr){6-9}
    Model     & AP$^\text{box}$  & AP$^\text{mask}$    & Params  & FLOPs  & mIoU$^\text{ss}$  & mIoU$^\text{ms}$    & Params  & FLOPs\\
    \midrule
    ConvNeXt-T\cite{liu2022convnet}&  50.4 &43.7 &86M & 741G  &46.0 &46.7& 60M& 939G\\
     Swin-T\cite{liu2021swin} &50.4& 43.7 &86M& 745G &44.5 &45.8 &60M& 945G\\
     
     HorNet-T\cite{rao2022hornet} &51.7 &44.8 &80M &730G&48.1 &48.9& 52M& 926G\\
   \rowcolor{mygray}\textbf{InfiNet-T} & 51.9  & 46.2& 77M & 724G  & 46.7  & 47.4& 50M  & 924G\\
    \midrule
        ConvNeXt-S\cite{liu2022convnet}&  51.9 &45.0& 108M &827G &48.7 &49.6 &82M& 1027G\\
     Swin-S\cite{liu2021swin} &51.8 &44.7& 107M &838G &47.6& 49.5& 81M& 1038G\\
     
     HorNet-S\cite{rao2022hornet} &52.7& 45.6 &107M &830G&49.2& 49.8 &81M& 1030G\\
  \rowcolor{mygray}\textbf{InfiNet-S}    & 52.8 & 46.4  & 98M&  802G & 49.4 & 49.9  & 78M&  1002G \\
   \midrule
       ConvNeXt-B\cite{liu2022convnet}& 52.7& 45.6& 146M &964G  &49.1 &49.9 &122M& 1170G\\
     Swin-B\cite{liu2021swin} &51.9 &45.0& 145M& 982G &48.1& 49.7& 121M& 1188G\\
     
     HorNet-B\cite{rao2022hornet} &53.3 &46.1& 144M &969G&50.0 &50.5 &121M& 1174G\\
 \rowcolor{mygray}\textbf{InfiNet-B}    & 53.7 &  47.3& 126M& 906G & 50.2 &  50.9& 105M& 1111G \\
  \midrule
      ConvNeXt-L$^\ddagger$\cite{liu2022convnet}&  54.8& 47.6 &255M & 1354G  &53.2 &53.7 &235M& 2458G\\
     Swin-L$^\ddagger$\cite{liu2021swin} &53.9& 46.7 &253M& 1382G &52.1 &53.5 &234M& 2468G\\
     
     HorNet-L$^\ddagger$\cite{rao2022hornet} &55.4& 48.0& 251M& 1363G&54.1 &54.5 &232M& 2473G\\
 \rowcolor{mygray}\textbf{InfiNet-XL$^\ddagger$} & 56.3  & 48.9  &  273M & 1454G& 54.6  & 55.2  &  253M & 2544G \\
    \bottomrule
  \end{tabular}
  }
\vspace{-3mm}
\end{table}

%% file: _tab/Ablation.tex
\begin{table}
  \caption{More Results on isotropic models and different kind of Reproducing Kernel}
  \label{tab:ablation}

  \begin{subtable}{.5\linewidth}
  \centering
  \tabcolsep 2pt
  \resizebox{.98\linewidth}{!}{
  \begin{tabular}{lllll}
    \toprule
     \multicolumn{4}{l}{(a)~Isotropic Models}\\
     \midrule
        &  Interact.  & Params& FLOPs& Top1 \\
      Model     & Orders  & (M)  & (G) & Acc.(\%) \\
    \midrule
    
    ConvNeXt-S(iso.) & no& 22 & 4.3 & 79.7\\
    Conv2Former(iso.) & 2 & 23 &4.3 & 81.2\\
        DeiT-S\cite{liu2022convnet} & 3 & 22  & 4.6 & 79.8\\
       HorNet-S(iso.)&2-5& 22 & 4.5  &80.6  \\
        \rowcolor{mygray} \textbf{InfiNet-S(iso.)}  &$\infty$  & 22   & 4.3 & 81.4 \\
   
    \bottomrule
  \end{tabular}}
  \end{subtable}
    \begin{subtable}{.5\linewidth}
  \centering
    \tabcolsep 2pt
    \resizebox{.98\linewidth}{!}{
  \begin{tabular}{lllll}
    \toprule
    \multicolumn{4}{l}{(b)~Ablation Study}\\
     \midrule
     &  Interact.  & Params& FLOPs& Top1 \\
   Model        & Orders  & (M)  & (G) & Acc.(\%) \\
    \midrule

    InfiNet-$\oplus$ & no& 23 & 3.2 & 81.6\\
       InfiNet-$\otimes$ & 2& 23 & 3.2 & 82.1\\
       InfiNet-2-polyno.& 4& 23 & 3.2  &82.3  \\
        InfiNet-3-polyno.& 6& 23 & 3.2  &82.5  \\
        \rowcolor{mygray} \textbf{InfiNet-T}  &$\infty$  & 23  & 3.2& 83.4 \\
    \bottomrule
  \end{tabular}}
  \end{subtable}
\vspace{-3mm}
\end{table}

%% file: _txt/6_Conclusion.tex
\section{Conclusion}

As a conclusion, in this paper, we propose that one of the key points of success of today's element-wise multiplication-based models is that they explore a high-dimensional feature interaction space through feature interactions. And the RBF kernel can greatly expand this interaction space into an infinite dimensional feature interaction space.  Based on this observation, we propose InfiNet, a high-performance neural network that explores infinite-dimensional feature interactions while using a modern model structure, which has achieved state-of-the-art results on several visual tasks.

\section*{Limitations}
\label{app:lim}
Although we propose the use of kernel methods for infinite-dimensional feature interaction, the only kernel methods we have tried so far are the RBF kernel function and some polynomial kernels of finite dimension. Substitutions utilizing a variety of kernel, including Laplace kernels, exponential kernels, a learnable kernel, etc., can be considered in subsequent studies. Our model has only been performed in some basic computer vision tasks, and validation in language and other modalities still requires some effort. The training of our model is only performed in the supervised learning paradigm, and more training and validation on self-supervised tasks still require effort.
In addition, to avoid additional hyperparameter tuning, we fixed the $\sigma$ parameter in the RBF kernel to 1. This may have deprived us of the possibility of exploring the optimal InfiNet, but due to the high cost of training the model, we will leave the impact of this hyperparameter on the InfiNet as a follow-up work.

\section*{Broader Social Impact}

\label{app:imp}
InfiNet is a state-of-the-art vision neural network architecture. The advancements in computer vision neural network architectures hold significant potential for positive societal impact, particularly in enhancing healthcare diagnostics, improving security systems, and advancing autonomous transportation. However, it is crucial to address potential negative implications such as privacy concerns, algorithmic biases, and job displacement. Ensuring ethical development and deployment involves implementing strict data protection measures, promoting fairness and inclusivity in algorithm design, and supporting workforce retraining programs. By proactively managing these challenges, we can maximize the benefits of computer vision technologies while minimizing their risks. Engaging with diverse stakeholders will be essential to ensure these technologies are used responsibly and equitably.

\section*{Acknowledgement}
We sincerely thank the reviewers for their valuable suggestions during the review stage.

%% file: _txt/8_Appendix.tex
\section{Appendix}
\subsection{Training Details}

\label{app:training}
The training details for ImageNet experiments are shown in Table~\ref{tab:in1k_config} and Table~\ref{tab:in21k_config}.
\input{_tab/moganetTabA2}
\input{_tab/moganetTabA3}

\newpage

\subsection{More Visualization Comparison}
\begin{figure}[t]
    \centering
    \includegraphics{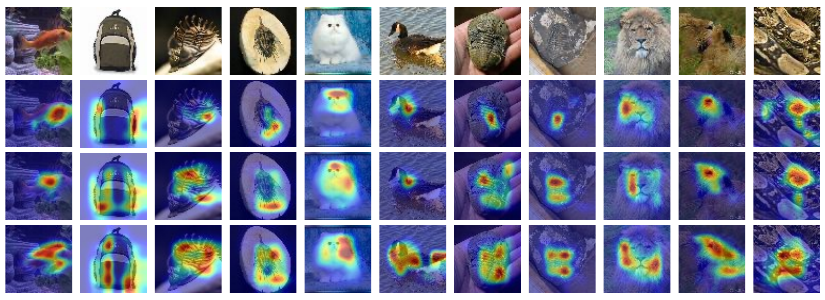}
    \includegraphics{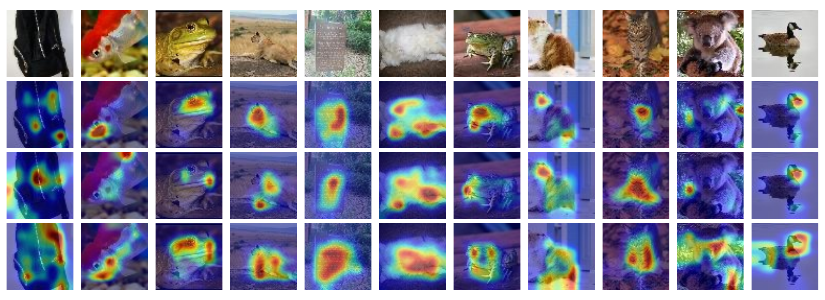}
    \includegraphics{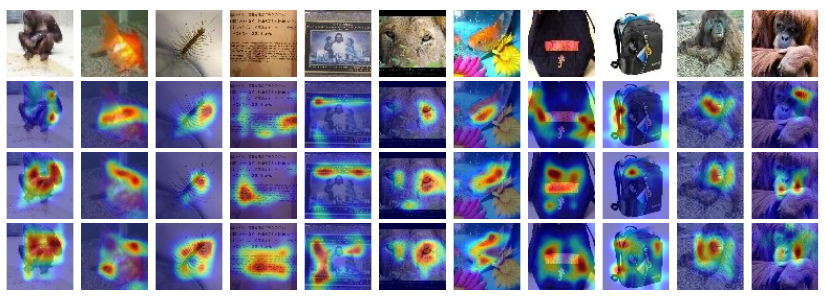}
    \caption{Visualization Comparison of (1) Feature Representation Space model, (2) Finite Feature Interaction Space model, (3) Infinite-Dimensional Feature Interaction model}
    \label{fig:4}
\end{figure} 
In Fig.~\ref{fig:4}, we provide more Class Activation Mapping to illustrate the interpretability of interaction models. Infinite-dimensional Interaction models capture more totality of objects within a category, illustrates the importance of this infinite-dimensional interaction.

%% file: _tab/moganetTabA2.tex
\begin{table}[h]
    \setlength{\tabcolsep}{5mm}
    \centering
    \caption{Training details for ImageNet-1K experiments}

\begin{tabular}{l|c}
    \toprule
    Configuration              &     InifNet-T/S/B/L            \\\hline
    Input resolution           & 224$^2$             \\
    Epochs                     & 300                 \\
    Batch size                 & 192/128/64/64                \\
    Optimizer                  & AdamW               \\
    AdamW $(\beta_1, \beta_2)$ & \small{$0.9, 0.999$}\\
    Learning rate              & 0.004               \\
    Learning rate decay        & Cosine              \\
    Weight decay               & 0.05                \\
    Warmup epochs              & 20                   \\
    Label smoothing $\epsilon$ & 0.1                 \\
    Stochastic Depth           & Y                   \\
    Rand Augment               & 9/0.5               \\
    Repeated Augment           & Y                   \\
    Erasing prob.              & 0.25                \\
    ColorJitter                & N                   \\
    Gradient Clipping          & Y                   \\
    EMA decay                  & Y                   \\
    \bottomrule
\end{tabular}

   
    \label{tab:in1k_config}
\end{table}

%% file: _tab/moganetTabA3.tex
\begin{table}[h]
    \setlength{\tabcolsep}{5mm}
    \centering
   \caption{Training details for ImageNet-21K experiments}
\begin{tabular}{l|cc|cc}
    \toprule
    Configuration              & \multicolumn{2}{c|}{IN-21K PT}              & \multicolumn{2}{c}{IN-1K FT}               \\ \cline{2-5}
                                    & L         & XL               & L        & XL      \\ \hline
    Input resolution           & \multicolumn{2}{c|}{224$^2$}                & \multicolumn{2}{c}{384$^2$}                \\
    Epochs                     & \multicolumn{2}{c|}{90}                     & \multicolumn{2}{c}{30}                     \\
    Batch size                 & \multicolumn{2}{c|}{256}                   & \multicolumn{2}{c}{64}                    \\
    Optimizer                  & \multicolumn{2}{c|}{AdamW}                  & \multicolumn{2}{c}{AdamW}                  \\
    AdamW $(\beta_1, \beta_2)$ & \multicolumn{2}{c|}{$0.9, 0.999$}           & \multicolumn{2}{c}{$0.9, 0.999$}           \\
    Learning rate              & \multicolumn{2}{c|}{$4\times 10^{-3}$}      & \multicolumn{2}{c}{$5\times 10^{-5}$}      \\
    Learning rate decay        & \multicolumn{2}{c|}{Cosine}                 & \multicolumn{2}{c}{Cosine}                 \\
    Weight decay               & \multicolumn{2}{c|}{0.1}                   & \multicolumn{2}{c}{0.00001}                   \\
    Warmup epochs              & \multicolumn{2}{c|}{5}                      & \multicolumn{2}{c}{0}                      \\
    Label smoothing $\epsilon$ & \multicolumn{2}{c|}{0.2}                    & 0.1             & 0.2     \\
    Rand Augment               & \multicolumn{2}{c|}{9/0.5}                  & \multicolumn{2}{c}{9/0.5}                  \\
    Repeated Augment           & \multicolumn{2}{c|}{N}                & \multicolumn{2}{c}{N}                               \\
    Erasing prob.              & \multicolumn{2}{c|}{0.25}                   & \multicolumn{2}{c}{0.25}                                    \\
    Gradient Clipping          & \multicolumn{2}{c|}{5}                & \multicolumn{2}{c}{5}                \\
    EMA decay                  & \multicolumn{2}{c|}{N}                & \multicolumn{2}{c}{Y}                \\
    \bottomrule
    \end{tabular}

    \label{tab:in21k_config}
\end{table}